# Investigating the effect of spontaneous speech features on by-speaker automatic speech recognition performance

Jessica Wormald and Lauren Harrington

*Department of Language and Linguistic Science, University of York, York, YO10 5DD, UK*

*jessica.wormald@york.ac.uk*, *lauren.harrington@york.ac.uk*

There is growing interest in using automatic speech recognition (ASR) systems to produce transcripts of real-life spoken interactions. ASR systems consistently perform less well with spontaneous speech compared with planned speech. Using a dataset of thirteen adult males of non-standard varieties of British English, we provide a linguistic evaluation of by-speaker ASR performance, analysing disfluency and articulatory imprecision to better understand ASR transcription errors. We observe between-speaker variability across the dataset and a complex interaction of speech features that can explain peripheral performance. The work demonstrates that ASR performance can be better understood by accounting for speech- and speaker-related factors.

## 1. Introduction

Most human speech is produced spontaneously, in the sense that it is not pre-planned or read from a script. There is growing interest in the use of AI-based speech technologies in the transcription of spoken interactions, particularly within high-stakes contexts such as healthcare[1] and policing[2]. Spontaneous speech is known to pose challenges for automatic speech recognition (ASR) systems[3], with consistently higher error rates for spontaneous speech in comparison with read speech[4]. Gabler et al.[5] discuss how spontaneous speech differs substantially from read speech as a result of its unplanned nature and note major characteristics that pose challenges for ASR systems, including disfluency patterns and phonetic reduction.

Disfluencies often feature in spontaneous speech[6]. Filled pauses (*uh* and *um*), word or phrase repetitions, and self-interruptions (e.g. *he was- he wore a bla- dark grey coat*) are commonly produced in unplanned speech, and have been shown to have a significantly negative impact on ASR performance for both non-pathological speakers[7] and speakers who stutter[8].

Reduction is also common in spontaneous speech, resulting in shortened or completely elided phonemes or syllables[9], as well as increased coarticulation and vowel centralisation[10]. This reduced articulatory precision results in less acoustic content for the ASR system to utilise when predicting what was said, and realisations of speech sounds that may not align with the acoustic characteristics associated with them by the ASR model. Spontaneous and unplanned speech also tends to be produced with a higher articulation rate than read speech and can also exhibit variability within an interaction[11]. Both very fast and very slow articulation rates have been shown to have a significant impact on ASR transcription performance[12].

The present study evaluates the impact of features associated with spontaneous speech on the performance of an open-source ASR model, CrisperWhisper[13]. Specifically, we explore disfluency rates and characterise articulatory precision by quantifying articulation rate and long-term vocal settings, to answer the following question: *How do features associated with spontaneous speech impact the outputs of ASR systems?*

We investigate variability in ASR performance within a demographic group, namely young male speakers of 'non-standard' varieties of British English. Our interest lies in the features of spontaneous speech, and individual speaker variability, that result in errors within the content of the ASR transcript. We also use a non-standard variety of British English given that accent, among other demographic factors, has been shown to negatively affect ASR performance[14,15,16]. We consider the implications of our findings for policy- and decision-makers in applied fields considering the use of ASR.

## 2. Methods

### 2.1 *Stimuli*

The stimuli for this study were extracted from the Punjabi English in Bradford and Leicester corpus[17], a database of spontaneous conversations between participants from either Bradford or Leicester and with either an Anglo or Punjabi language background (i.e. parent(s) or grandparent(s) from the Punjab region). We used a small subset of thirteen young male speakers in the corpus who had consented to their data being used in future research. There were two Anglo speakers from Bradford (Jake, Matthew), two Anglo speakers from Leicester (Andy, Simon), three Punjabi speakers from Bradford (Halim, Hamza, Irfan), and six Punjabi speakers from Leicester (Adeep, Hanspal, Hardev, Jasbir, Nihal, Satbir).

In order to explore spontaneous speech features in depth, samples were edited to contain the voice of a single speaker and limited to five minutes in duration (between 301 and 310 seconds). Topics of conversation included hobbies, religion and hometowns. Participants were recorded in a quiet room with neck-worn microphones, thus representing good quality audio recordings and favourable conditions for the ASR system.

### 2.2 *ASR transcription & evaluation*

Transcripts were generated using CrisperWhisper[13], a variant of OpenAI's state-of-the-art Whisper model[18], which claims to transcribe disfluencies and produce time stamps with higher accuracy. JSON outputs were converted to Praat TextGrids[19], and a reference transcript for each sample was produced by correcting the CrisperWhisper output via several rounds of review by multiple analysts. Transcripts were aligned on a word-level basis and error types were determined according to whether the word(s) in the reference and ASR transcripts matched or constituted an error. Errors were categorised as either substitutions (words in the transcripts differed), deletions (no interpretation in the ASR transcript) or insertions (additional content in the ASR transcript). Performance was evaluated by calculating (i) the number of each error type produced per speaker, and (ii) the word error rate (WER) per speaker. WER was calculated by dividing the total number of errors by the total number of words in the reference transcript. The stimuli and error quantification employed in this study also feature in Harrington *et al.*[20], which explores in detail the errors found within the ASR transcripts.

### 2.3 *Speech feature analysis*

We focus our linguistic analyses on disfluency production and articulatory precision, characterised by articulation rate and long-term vocal settings. Disfluency analysis was based on the Taxonomy of Fluency Features for Forensic Analysis[21]. We focused our analysis on interruptions (word and phrase), repetitions (part-word, word and phrase) and filled pauses (*uh* and *um*). We expect that these disfluencies are most likely to impact the content of ASR transcripts given that they affect the predictability of the speech at both the word and phrase level. We combine part-word repetitions, word repetitions and word interruptions into a 'word-level' category, and phrase repetitions and phrase interruption into a 'phrase-level' category. Disfluencies were marked in Praat TextGrids to facilitate the calculation of overall counts as well as detailed examination of what types of errors occurred during disfluent sections.

To characterise articulatory precision, we measured (i) articulation rate, and (ii) the long term settings of vocal tract tension and lingual posture. To calculate articulation rate, we used pauses of over 250 milliseconds[22] to demarcate utterances (i.e. inter-pause stretches), and we quantified articulation rate using phonological syllables within an utterance, in line with Gold[23]. By choosing to measure articulation rate by phonological syllables, we will likely see higher rates for speakers who reduce or elide syllables in running speech[24]. Thus, we have also measured overall vocal tract tension alongside a general assessment of lingual posture.

Vocal tract tension and lingual posture are long-term vocal settings which are considered relative to a 'neutral' posture[25]. Lax vocal tract refers to a general undershoot or reduced articulatory effort throughout speech; segments are generally shorter with less movement of the articulators and attenuation of acoustic information is common[25]. Lingual posture refers to the long term position of the tongue, which may be fronted or retracted[26]. For each setting, speakers were characterised along separate mutually exclusive scales by the first author in line with San Segundo and Mompean[27]. We included two scalar degrees of variability from neutral in either direction as we are particularly interested in the between-speaker variability associated with these settings, and are not completing a full vocal profile analysis.

## 3. Results

### 3.1 *Overall performance*

Average performance for CrisperWhisper with this dataset is 9.88% WER. We see widespread variability from the best transcript to the worst, with total error counts ranging from 34 to 399 errors across a duration of 301-310 seconds. As shown in Figure 1, by-speaker WER ranges from 3.06% to 32.39%, though for nine of the thirteen speakers, performance was at a similar level: around 50-80 errors and 6-8.5% WER. These error rates are relatively consistent with those previously reported for CrisperWhisper[13] (3.26-9.72% WER). For three speakers, we observed overall poor performance. The transcripts for Jake and Andy each contained 150-170 errors and WERs surpassed 14%, and performance for Matthew was considerably worse, with almost 400 transcription errors and a WER of over 30%. In addition to overall by-speaker WERs, Figure 1 highlights how errors are distributed across the three error types (substitutions, deletions and insertions).

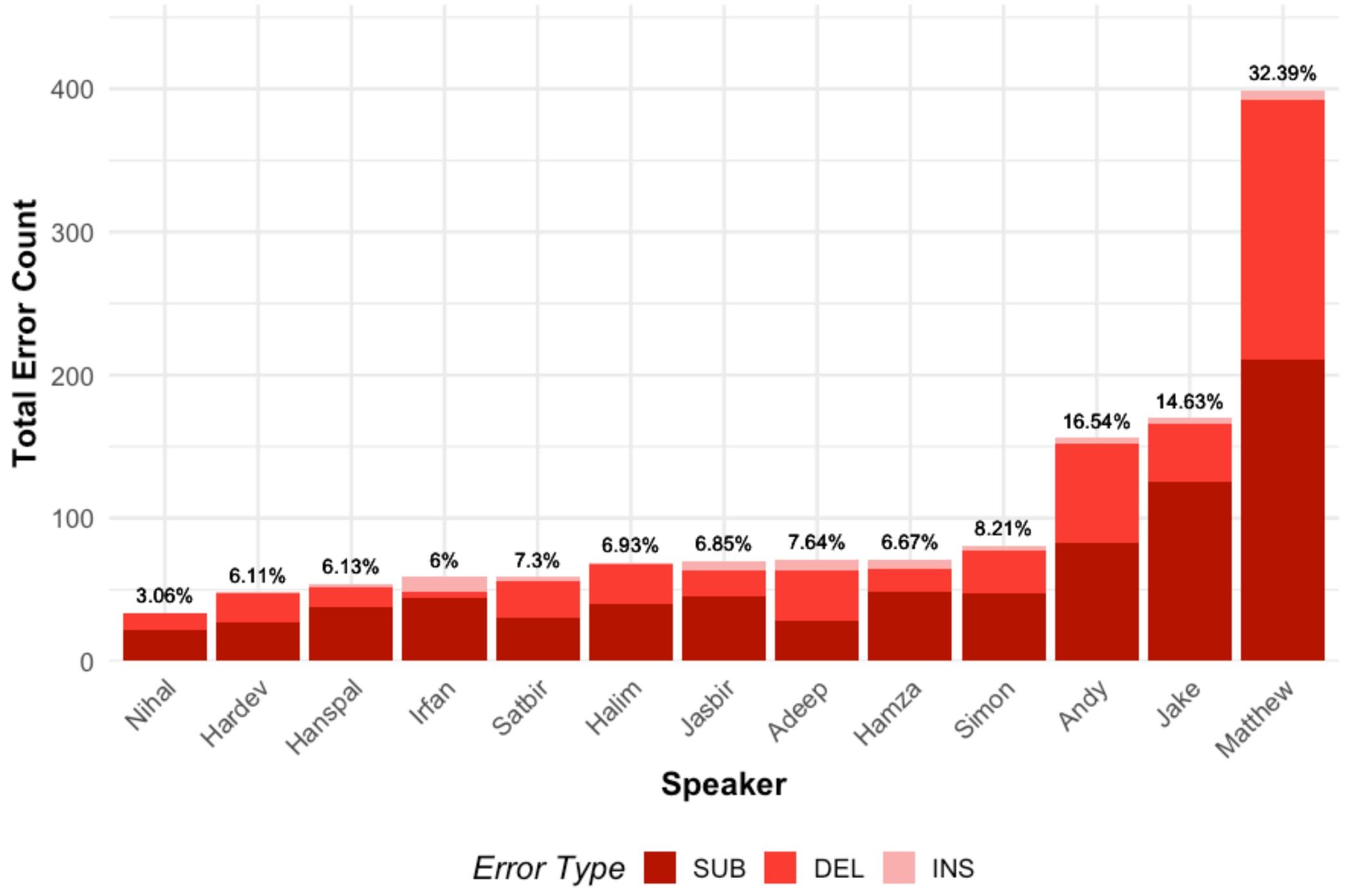

Fig. 1: Total error counts per speaker, ordered from lowest to highest total number of errors. The bars are colour-coded according to the three error types (substitutions, deletions and insertions). WER appears in bold text above each speaker bar.

### 3.2 *Disfluencies*

Figure 2 shows the speakers, ordered by increasing total error counts, and their (a) disfluency production, (b) articulation rate, and (c) long-term vocal settings associated with vocal tract laxness and fronted lingual posture. As highlighted in Figure 2a, there was variability in the total disfluency counts and types most commonly produced across speakers. The total number of disfluencies (interruptions, repetitions and filled pauses) varied across the 5-minute samples for the speakers, from 31 (Simon) to 111 (Adeep). Filled pauses (*uh* and *um*) were the most commonly produced disfluency type for eight of the thirteen speakers and word-level disfluencies were most common for the remaining five, including the two speakers for whom the ASR system produced the most errors: Jake and Matthew.

A high number of disfluencies did not routinely result in more errors within the ASR transcripts. For example, Adeep produced the highest number of disfluencies but had a relatively average error count. However, most of his disfluencies were filled pauses, and we found that 93% of filled pauses within the dataset were accurately transcribed by CrisperWhisper. Accuracy was substantially lower for phrase-level disfluencies (73-78%) and lowest for word-level disfluencies (30-56%). In contrast to Adeep, the speaker with the second highest disfluency count (Matthew) produced mostly word-level disfluencies, and he had the highest number of transcription errors. However, it should be noted that although word-level disfluencies were transcribed with lower accuracy, we did not find a consistent pattern regarding poor performance and word-level disfluencies. Another speaker (Halim) produced a similar number of word-level disfluencies as Matthew, yet his ASR transcript contained a relatively average number of errors.

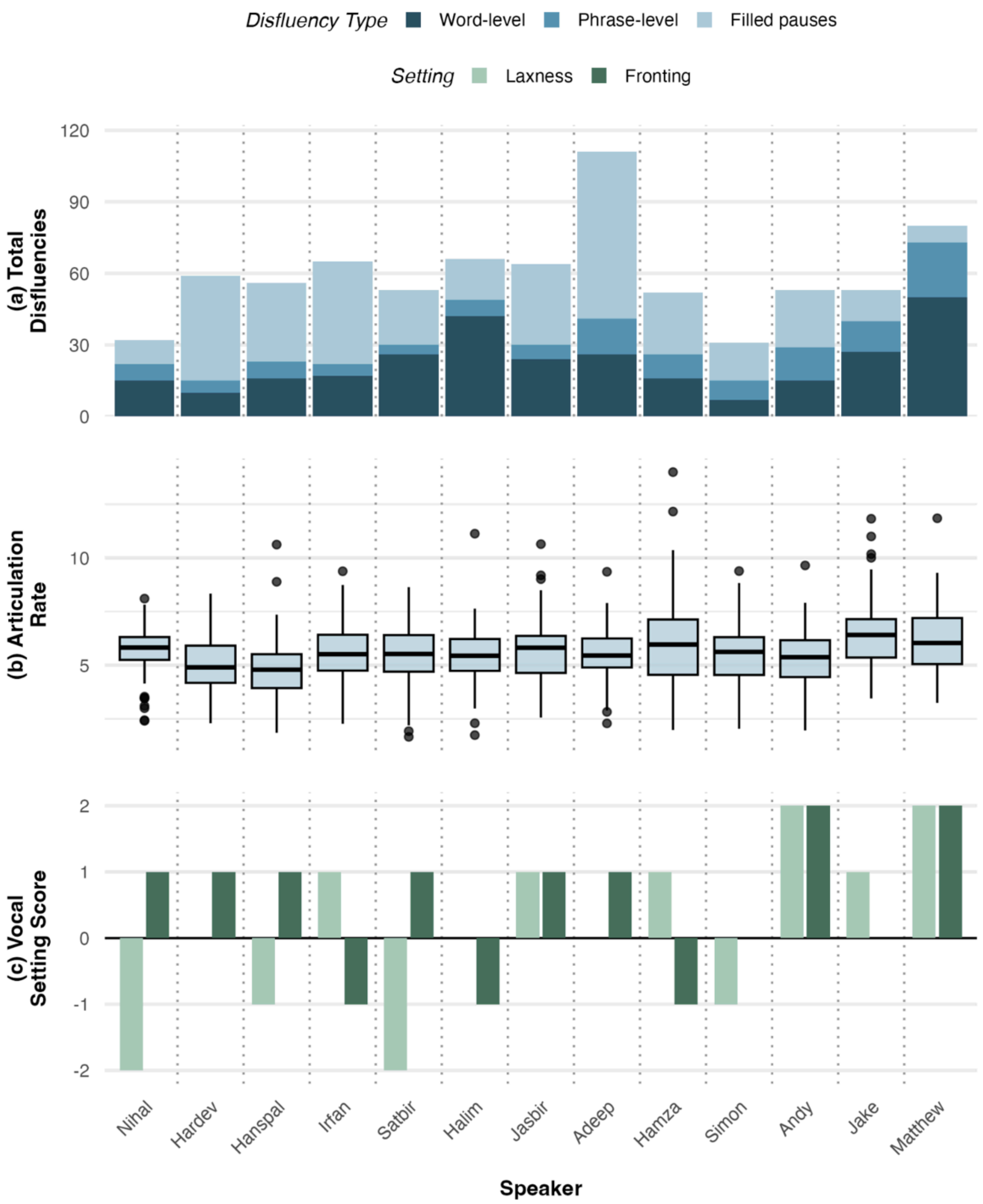

Fig. 2: By-speaker (a) disfluency production, (b) articulation rate and (c) long-term vocal settings related to vocal tract tension and lingual posture. Disfluencies are divided into word-level (part-word repetitions, and word interruption and repetitions), phrase-level (phrase interruptions and repetitions), and filled pauses. Articulation rates for all utterances are included. Positive vocal setting scores

correspond to increased laxness (light green) and fronting (dark green). Speakers are ordered from lowest to highest number of errors, as in Fig. 1.

### 3.3 *Articulatory precision*

Articulatory precision was characterised by articulation rate (Figure 2b), and long-term vocal settings (Figure 2c). Average by-speaker articulation rates ranged from 4.8 syllables per second (Hanspal) to 6.4 syllables per second (Jake). This range is at the lower end of what has previously been found for male speakers of Southern Standard British English[23]. The two speakers with the highest number of ASR errors are also the only speakers with an articulation rate over 6 syllables per second: Jake at 6.4, and Matthew at 6.0. However, there is no consistent relationship between articulation rate and ASR performance. Nihal, for whom the ASR system performed best, had a fairly high articulation rate within the group (5.7 syllables per second), and Hanspal, for whom the ASR system also performed well, had the lowest articulation rate in our dataset (4.8 syllables per second).

When investigating vocal settings, we found considerable variability between the speakers. Light green bars in Figure 2c represent vocal tract tension and dark green bars represent lingual posture. A score of +2 corresponds to a markedly lax vocal tract or markedly fronted lingual posture, respectively. Some degree of lingual fronting was common for most speakers, and backed lingual posture was only observed for Bradford Punjabi males (see Wormald[17] for a more comprehensive discussion of the voice quality patterns in this dataset). With regard to overall vocal tract tension, most speakers are around neutral or with only slight tension or laxness. Nihal and Satbir are the only speakers with a markedly tense vocal tract, i.e. their articulations are very precise and clear, and we observe relatively low error counts for these speakers. In contrast, the two speakers with a markedly lax vocal tract (Andy and Matthew) have two of the highest overall error counts in the dataset.

However, in addition to vocal tract laxness, Andy and Matthew are also the only speakers to have a markedly fronted lingual posture. We found several examples for both speakers of tongue-tip articulations being made with the tongue-blade, and this fronting was much more extreme than the general pattern of fronting noted for many speakers. This reduced articulatory precision results in high numbers of errors which are principally substitutions and deletions, whereby the ASR system has either misinterpreted or simply omitted the speech content.

When considered independently, there is no consistent pattern for any of the speech-related features and their impact on ASR performance. We do not observe a systematic increase in any of these features as total error count increases. We do, however, see that for speakers at each end of our error count distribution there is a clearer relationship between at least one of the speech features and ASR performance. We argue that it is a combination of speech factors which helps us understand the error counts, but not systematically predict ASR performance.

## 4. Discussion

When considering overall performance, CrisperWhisper[13] produces generally accurate transcriptions, with the average error rate at 9.88% for this dataset of young male speakers of 'non-standard' varieties of British English. All transcripts contain errors owing to the unpredictable, messy and variable nature of spontaneous speech. Without an understanding of the possible parameters of this variation, it is difficult to predict how a system will perform with a given voice. This finding is echoed in studies on automatic speaker recognition systems which explore individual speaker behaviour and variability in voices[28,29]. These studies demonstrate that some speakers can impact automatic speaker recognition performance in a non-generalisable manner.

Features of spontaneous speech do not impact ASR performance in a linear or consistently predictable way. For disfluency, sub-word units are more likely to be inaccurately transcribed. Filled pauses are generally accurately transcribed by CrisperWhisper, and phrase-level disfluencies are somewhere in between. Thus, whether a speaker is disfluent in general does not necessarily predict that the system will struggle. Instead, it is specifically how they are disfluent which is important. In our dataset, Adeep produces the most disfluencies but these are principally filled pauses, and the ASR transcript of his speech contains a relatively average number of errors compared to others in our dataset. In contrast, Matthew produces fewer disfluencies overall but these mostly involve sub-word units, and his ASR transcript contains over five times more errors than Adeep's.

With regard to articulatory precision, this was assessed by considering articulation rate and vocal settings. The system performs best with Nihal and worst with Matthew. These speakers have somewhat similar articulation rates but considerable differences in vocal tract tension and overall disfluency rates. Matthew has a markedly lax vocal tract and is very disfluent: phonetic content is often reduced or removed, articulatory targets are not being met, and disfluencies are principally at the word-level. As noted earlier, in contexts such as these there is less content for the ASR system to use to predict the words spoken. The underarticulation means that the acoustic information that is present is not representative of canonical articulations. This is compounded by the disfluency pattern; content is often uninterrupted which inhibits the system predicting what might come next. In contrast, Nihal, the speaker with the lowest number of errors, has a markedly tense vocal tract. Although he speaks quickly, his speech is clearly articulated (i.e. he meets articulatory targets) and utterances are not often broken up with word interruptions. When speakers increase their speech rate there are two ways they can do this: (i) maintain closures and distinct articulations, just make these more quickly, as with Nihal, or (ii) reduce the number of clearly articulated segments via elisions, reductions and simplifications, as with Matthew. Thus, for Nihal, the ASR system receives more acoustic information about articulatory patterns and there are few incomplete utterances, meaning the system can more reliably predict what follows.

When we consider these features in combination, we can more comprehensively understand system performance. Matthew, the speaker with worst overall performance, has (i) one of the highest disfluency counts and disfluencies that principally involve sub-word units, (ii) an extremely lax vocal tract, (iii) markedly front lingual posture, and (iv) one of the highest articulation rates. We believe this combination of speech features are substantially contributing to the challenges the ASR system is having when transcribing Matthew's speech. Matthew is an outlier, but the poorer ASR performance is completely predictable. His speech represents the limits of what the ASR system can reasonably be expected to do. Here, the errors do not necessarily highlight areas where the system could be improved. Instead, the speech-related factors add context and facilitate better understanding of system performance.

Previous research in this area highlights that demographic factors have an impact on ASR performance[14,15,16]. In our data, we found that the ASR system performed better for Punjabi English speakers than Anglo English speakers. However, our dataset only includes four Anglo English speakers, and for two of them (Andy and Matthew), we believe the combination of speech-related factors observed are a significant contributor to the decline in ASR performance. We do, however, believe that regional accent features account for the high error rate we observe for Jake from Bradford. Jake's errors are largely substitutions and these are often related to accent features, such as definite article reduction (e.g. "sit in [the] park") and monophthongal vowel realisations, similar to other findings noted for speakers of West Yorkshire English[16]. This research highlights that demographic group-level characteristics alone do not fully explain the potential variability in performance of ASR systems. We have demonstrated that the speech itself can also be challenging as a result of a complex interaction between different factors.

## 5. Conclusions

In this paper we have explored how features associated with spontaneous speech impact the outputs of an ASR system. We have demonstrated a complex pattern of interacting features and that there is no clear correspondence between error counts and one particular speech feature. Disfluencies at the word-level can result in more ASR errors than filled pauses or phrase-level disfluencies. Reduced articulatory precision related to vocal tract laxness and marked lingual fronting can also impact on the acoustic content available to the ASR system and lead to increased error rates. When these features combine, we see effects of a greater magnitude. These findings provide important context for those wanting to adopt ASR into organisational workflows, particularly in high stakes contexts, highlighting critical factors which influence ASR performance but may not be as readily considered by decision-makers as demographic or technical factors.

## Acknowledgments

This work was supported by an International Association for Forensic and Acoustic Phonetics research grant. We acknowledge the contributions of Tallulah Buckley with data analysis, the University of York's Viking high compute cluster and Philip Harrison for computational processing and support, and Vincent Hughes and James Tompkinson for reviewing drafts of this paper.

## Author Declarations

The authors have no conflicts of interests to disclose.

## Data Availability

The data that support the findings of this study are available from the corresponding author upon reasonable request.